\def\year{2021}\relax
\documentclass[letterpaper]{article} 
\usepackage{21}  
\usepackage{times}  
\usepackage{helvet} 
\usepackage{courier}  
\usepackage[hyphens]{url}  
\usepackage{graphicx} 
\usepackage{footnote}
\usepackage{enumitem}
\usepackage{subfigure}
\usepackage[switch]{lineno}
\newcommand{\mb}{\mathbf}
\usepackage{natbib}  
\usepackage{caption} 
\title{Contrastive Self-supervised Learning for Graph Classification}
\author {
    Jiaqi Zeng,
    Pengtao Xie
}

\affiliations {
    \\UC San Diego \\
     pengtaoxie2008@gmail.com
}

\begin{document}

\maketitle
\begin{abstract}
Graph classification is a widely studied problem and has broad applications. In many real-world problems, the number of labeled graphs available for training classification models is limited, which renders these models prone to overfitting. To address this problem, we propose two approaches based on contrastive self-supervised learning (CSSL) to alleviate overfitting. In the first approach, we use CSSL to pretrain graph encoders on widely-available unlabeled graphs without relying on human-provided labels, then finetune the pretrained encoders on labeled graphs. In the second approach, we develop a regularizer based on  CSSL, and solve the supervised classification task and the unsupervised CSSL task simultaneously. To perform CSSL on graphs, given a collection of original graphs, we perform data augmentation to create augmented graphs out of the original graphs. An augmented graph is created by consecutively applying a sequence of graph alteration operations. A contrastive loss is defined to learn graph encoders by judging whether two augmented graphs are from the same original graph. Experiments on various graph classification datasets demonstrate the effectiveness of our proposed methods.
\end{abstract}

\section{Introduction}
Graph classification~\cite{zhang2019hierarchical,di2019mutual} is a widely studied problem in machine learning and data mining and finds broad applications. For example, given a molecule graph of a protein, judge whether this protein is non-enzyme. Given a chemical compound graph, judge whether the compound is mutagen or non-mutagen. In many real-world graph classification problems, the number of graphs available for training is oftentimes limited. For instance, it is difficult to obtain a lot of protein graphs in many biomedical studies due to the financial cost. It is well known that when the amount of training data is limited, the model tends to overfit to the training data and perform less well on test data.

To address the overfitting problem in graph classification, we propose two approaches: CSSL-Pretrain and CSSL-Reg, both based on contrastive self-supervised learning (CSSL)~\cite{he2019moco,chen2020simple,chen2020mocov2}.
In CSSL-Pretrain, we use CSSL to pretrain graph encoders on widely-available unlabeled graphs without relying on human-provided labels, then finetune the pretrained encoders on labeled graphs. In CSSL-Reg, we develop a regularizer based on  CSSL, and solve the supervised classification task and the unsupervised CSSL task simultaneously. Self-supervised learning (SSL)~\cite{gidaris2018unsupervised,pathak2016context,zhang2016colorful} is an unsupervised learning approach which defines auxiliary tasks on input data without using any human-provided labels and learns data representations by solving these auxiliary tasks. Contrastive SSL~\cite{he2019moco,chen2020mocov2,chen2020simple} creates augmentations of original data examples and defines an auxiliary task which judges whether two augmented data examples originate from the same original data example.  Recently, several self-supervised learning approaches~\cite{peng2020self,qiu2020gcc,sun2019infograph} are proposed for representation learning on graphs. These approaches focus on learning representations of local elements in graphs, such as nodes and subgraphs. In contrast, our method focuses on learning graph-level representations that are more suitable for tasks like graph classification.

To perform CSSL on graphs, we first create augmented graphs from the original graphs, based on four basic graph alteration operations including edge deletion, edge insertion, node deletion, and node insertion. To create an augmented graph, we apply a sequence of graph alteration operations consecutively: the operation at step $t$ is applied to the intermediate graph generated after applying the operation at step $t-1$. Given the augmented graphs, we define a CSSL task to distinguish whether two augmented graphs are created from the same original graph. In CSSL-Pretrain, we first pretrain a graph encoder by solving the graph CSSL task, then use this pretrained encoder as initialization and continue to finetune it by minimizing the graph classification loss. CSSL-Pretrain learns powerful graph representations on unlabeled graphs (which are widely available) in an unsupervised way without relying on human-provided labels. Since these representations are learned without using labels, they are less likely to be overfitted to the labels in the small-sized training dataset and hence help to reduce overfitting. In CSSL-Reg, the CSSL loss serves as a regularization term and is optimized jointly with the classification loss.    CSSL-Reg enforces the graph encoder to jointly solve two tasks: an unsupervised CSSL task and a supervised graph classification task. Due to the presence of the CSSL task, the model is less likely to be biased to the classification task defined on the small-sized training data.
We perform experiments on five datasets. Our proposed CSSL-Pretrain and CSSL-Reg outperform baseline approaches, which demonstrate the effectiveness of our methods in alleviating overfitting.

The major contributions of this paper are as follows:
\begin{itemize}[leftmargin=*]
    \item We propose CSSL-Pretrain, which is an unsupervised pretraining method of graph encoders based on contrastive self-supervised learning, to learn graph representations that are resilient to overfitting.
    \item We propose CSSL-Reg, which is a data-dependent regularizer based on CSSL, to reduce the risk that the graph encoder is biased to the data-deficient classification task on the small-sized training data.
    \item Experiments on various datasets demonstrate the effectiveness of our approaches.
\end{itemize}

The rest of the paper is organized as follows. Section 2 reviews related works. Section 3 and 4 present the methods and experiments respectively. Section 5 concludes the paper and discusses future works.

\section{Related Works}

\subsection{Graph Representation Learning}
In graph applications, learning useful representations of nodes, edges, and the entire graph is crucial for downstream applications such as graph classification, node classification, graph completion, etc.
Classic approaches for graph representation learning can be categorized as: (1)  embedding methods: for example, DeepWalk \cite{perozzi2014deepwalk} leveraged truncated random walk to learn node embeddings, LINE \cite{tang2015line} used edge sampling to learn node embeddings in large-scale graphs, HARP \cite{chen2017harp} utilized hierarchical representation learning to capture global structures in graphs; (2) matrix-factorization-based methods: for example, NetMF \cite{qiu2018network} discovered a theoretical connection between DeepWalk’s implicit matrix and graph Laplacians and proposed an embedding approach based on this connection, HOPE \cite{ou2016asymmetric} proposed an asymmetric transitivity preserving graph representation learning method for directed graphs. 

Recently, graph neural networks (GNNs) have achieved remarkable performance for graph modeling.
GNN-based approaches can be classified into two categories: spectral approaches and message-passing approaches. The spectral approaches generally use graph spectral theory to design parameterized filters. Based on Fourier transform on graphs, \citet{bruna2013spectral} defined convolution operations for graphs. To reduce the heavy computational cost of graph convolution, \citet{defferrard2016convolutional} utilized fast localized spectral filtering. Graph convolution network (GCN) \cite{kipf2016semi}  truncated the Chebyshev polynomial to the first-order approximation of the localized spectral filters. The message-passing approaches basically aggregate the neighbours' information through convolution operations. GAT \cite{velivckovic2017graph} leveraged attention mechanisms to aggregate the neighbours' information with different weights. GraphSAGE \cite{hamilton2017inductive} generalized representation learning to unseen nodes using neighbours' information. Graph pooling methods such as DiffPool \cite{ying2018hierarchical} and HGP-SL \cite{zhang2019hierarchical} were developed to aggregate node-level representations into graph-level representations.


\subsection{Contrastive Self-supervised Learning}
Contrastive self-supervised learning has arisen much research interest recently. MoCo~\cite{he2019moco}  and SimCLR~\cite{chen2020simple} learned image encoders by predicting whether two augmented images were created from the same original image. \citet{henaff2019data} studied data-efficient image recognition based on  contrastive predictive coding~\cite{oord2018representation}, which  predicted the future in latent space by using powerful autoregressive models. \citet{srinivas2020curl} proposed to learn  contrastive unsupervised representations for reinforcement learning. \citet{khosla2020supervised} investigated supervised contrastive learning, where clusters of points belonging to the same class were pulled together in embedding space, while clusters of samples from different classes were pushed apart. \citet{klein2020contrastive} proposed a contrastive self-supervised learning approach for commonsense reasoning. \citet{he2020sample} proposed a Self-Trans approach which applied contrastive self-supervised learning on top of networks pretrained by transfer learning.

\subsection{Self-supervised Learning on Graphs}
Recently, several self-supervised learning approaches are proposed for representation learning on graphs.
\citet{peng2020self} learned node representations by randomly selecting pairs of nodes in a graph and training a neural net to predict the contextual position of one node relative to the other. GCC~\cite{qiu2020gcc} defined the pre-training task as subgraph instance discrimination in and across networks and leveraged contrastive learning to learn structural representations. InfoGraph \cite{sun2019infograph} defined SSL tasks which maximize mutual information between graph representations and sub-structural representations. These approaches focused on learning representations of local elements in graphs, such as nodes and subgraphs. In contrast, our method focuses on learning graph-level representations that are more suitable for tasks like graph classification.

\section{Methods}
To alleviate overfitting in graph classification, we propose two methods based on contrastive self-supervised learning (CSSL): CSSL-Pretrain and CSSL-Reg. In CSSL-Pretrain, we use CSSL to pretrain the graph encoder. In CSSL-Reg, we use the CSSL task to regularize the graph encoder. 

\subsection{Contrastive Self-supervised Learning on Graphs}
In this section, we discuss how to perform contrastive self-supervised learning on graphs, which is the basis of CSSL-Pretrain and CSSL-Reg. Self-supervised learning (SSL)~\cite{gidaris2018unsupervised,pathak2016context,zhang2016colorful} is a learning paradigm that aims to capture the intrinsic patterns and properties of input data  without using human-provided labels.
The basic idea of SSL is to construct some auxiliary tasks solely based on the input data itself without using human-annotated labels and make the network to learn meaningful representations by performing the auxiliary tasks well, such as  rotation prediction~\cite{gidaris2018unsupervised},  image inpainting~\cite{pathak2016context},  automatic colorization~\cite{zhang2016colorful}, context prediction~\cite{nathan2018improvements}, etc.
The auxiliary tasks in SSL can be constructed using many different mechanisms. Recently, a contrastive mechanism~\cite{hadsell2006dimensionality} has gained increasing attention and demonstrated promising results in several studies~\cite{he2019moco,chen2020mocov2}. The basic idea of contrastive SSL is: generate augmented examples of original data examples, create a predictive task that predicts whether two augmented examples are from the same original data example or not, and learn the representation network by solving this task.
\begin{figure}[t]
        \centering
        \quad
        \subfigure[]{
        \label{fig:origin}
        \includegraphics[width=0.1\textwidth]{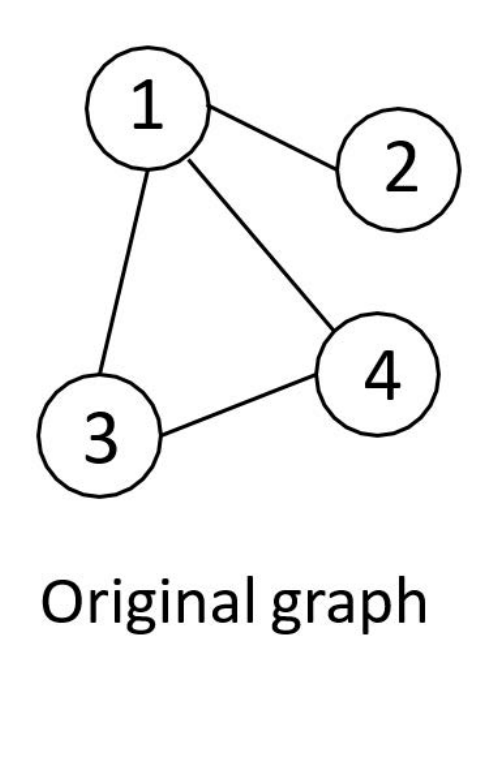}
        }
        \quad
        \subfigure[]{
        \label{fig:edge_del}
        \includegraphics[width=0.1\textwidth]{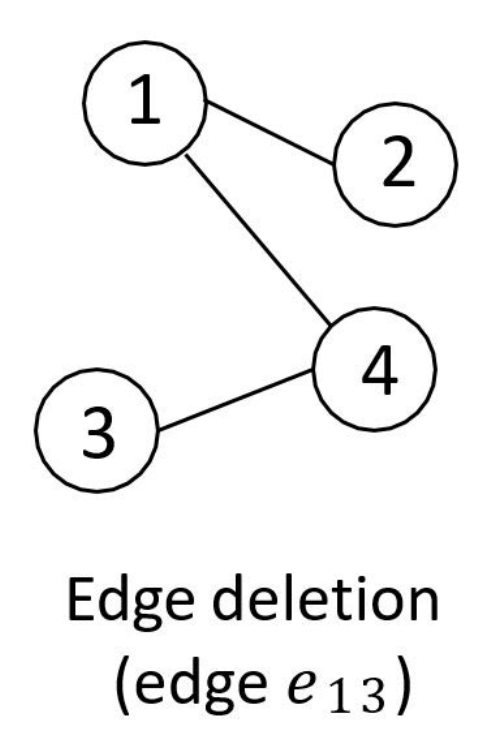}
        }
        \quad
        \subfigure[]{
        \label{fig:node_del}
        \includegraphics[width=0.1\textwidth]{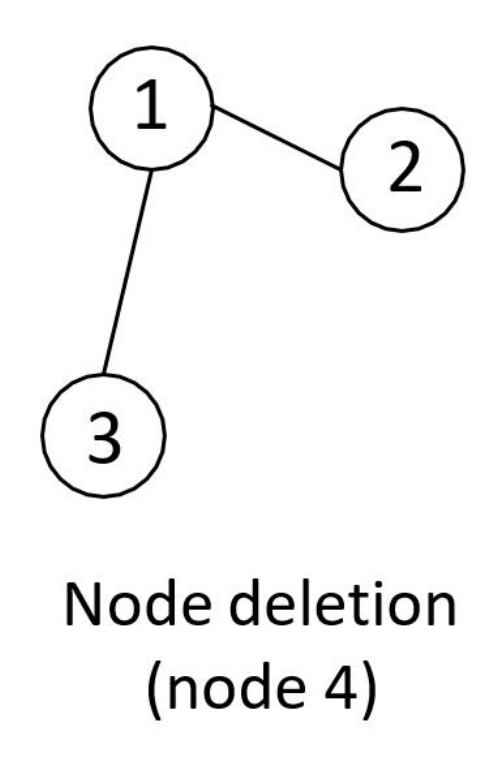}
        }
        \quad
        \subfigure[]{
        \label{fig:edge_add}
        \includegraphics[width=0.1\textwidth]{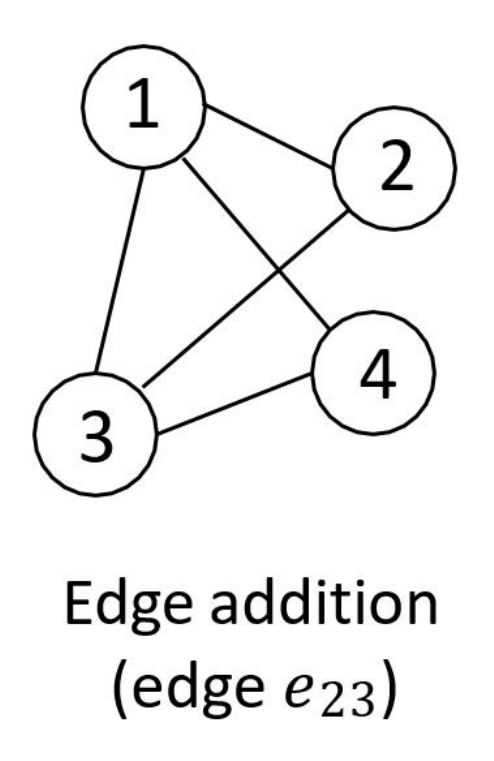}
        }
        \quad
        \subfigure[]{
        \label{fig:node_add}
        \includegraphics[width=0.1\textwidth]{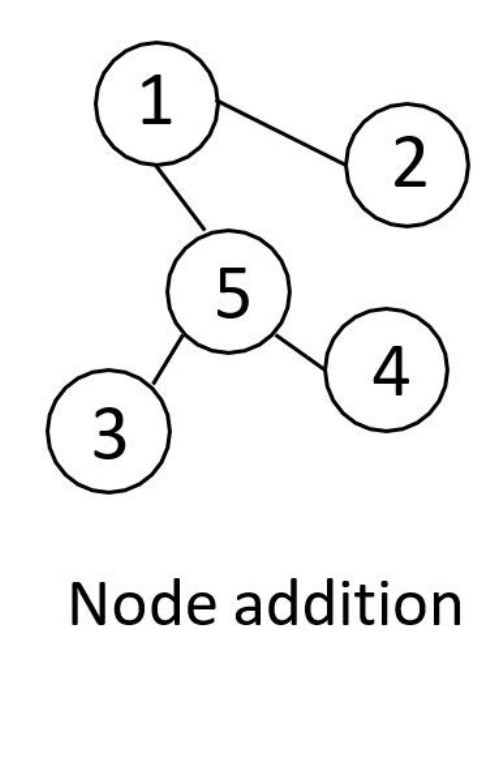}
        }
        \caption{Graph alteration operations.}
        \label{fig:gao}
  \end{figure}

To perform CSSL on graphs, given a collection of original graphs, we perform graph augmentation to generate augmented graphs from the original graphs, then learn a network to predict whether two augmented graphs originate from the same original graph or not. To perform graph augmentation, we use four types of basic graph alteration operations, as illustrated in Figure~\ref{fig:gao}. The four types of operations include:
\begin{itemize}
    \item \textbf{Edge deletion}: randomly select an edge and remove it from the graph. For example,
   in Figure \ref{fig:edge_del}, we randomly select an edge (which is the one between node 1 and 3), and delete it.
    \item \textbf{Node deletion}: randomly select a node and remove it from the graph; remove all edges connecting to this node. For example,
   in Figure \ref{fig:node_del}, we randomly select a node (which is 4), delete this node and all edges connected with node 4.
   \item \textbf{Edge insertion}: randomly select two nodes, if they are not directly connected but there is a path between them, add an edge between these two nodes. For example,
   in Figure \ref{fig:edge_add}, node 2 and 3 are not directly connected, but there is a path between them ($2\to1\to 3$). We connect these two nodes with an edge.
   \item \textbf{Node insertion}: randomly select a strongly-connected subgraph $S$, remove all edges in $S$, add a node $n$, and add an edge between $n$ and each node in $S$. For example,
   in Figure \ref{fig:node_add}, node 1, 3, 4 form a complete subgraph. We insert a new node 5, connect node 1, 3, 4 to node 5, and remove the edges among node 1, 3, 4.
\end{itemize}
Given an original graph $G$, to create an augmentation of $G$, we apply a sequence of graph alteration operations consecutively. At step 1, we randomly sample an operation $o_1(\cdot)$ that is applicable to $G$, perform this operation and get an altered graph $G_1=o_1(G)$. At step 2, we randomly sample another operation $o_2(\cdot)$ that is applicable to $G_1$, perform this operation and get $G_2=o_2(G_1)$. This procedure continues until the maximum number of steps is reached. At each step $t$, an applicable operation is randomly sampled and applied to the intermediate graph $G_{t-1}$ generated at step $t-1$.


\begin{figure}[t]
    \centering
    \includegraphics[width=0.5\textwidth]{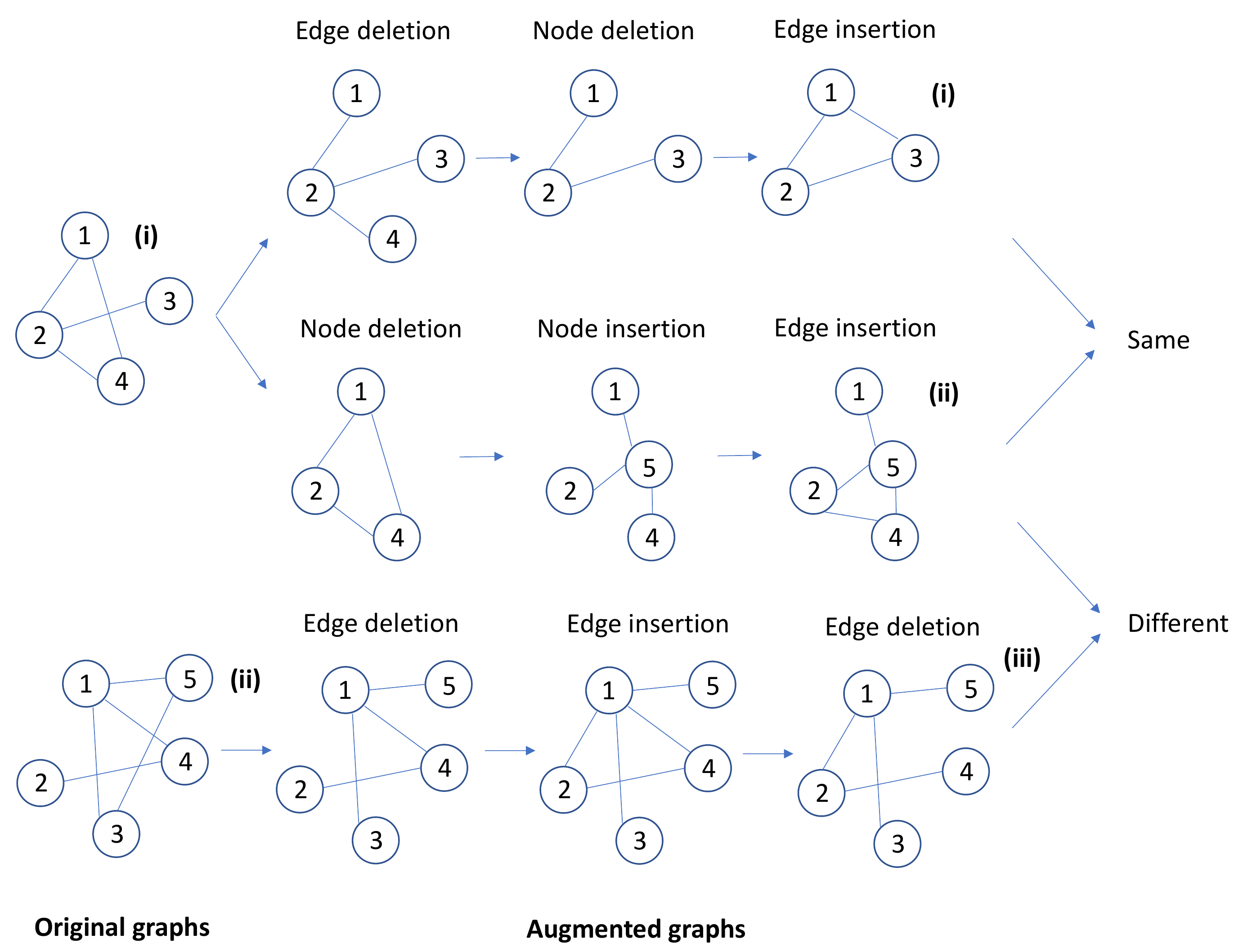}
    \caption{ Illustration of contrastive SSL on graphs.
    }
    \label{fig:illustration}
\end{figure}

Next, we define the contrastive learning loss on augmented graphs. If two augmented graphs are created from the same original graph, they are labeled as being similar; otherwise, they are labeled as dissimilar. We learn a network to fit these similar/dissimilar binary labels. The network consists of two modules: a graph embedding module $f(\cdot)$ which extracts the latent representation $\mathbf{h}=f(\mathbf{x})$ of a graph $\mathbf{x}$  and a multi-layer perceptron $g(\cdot)$ which takes $\mathbf{h}$ as input and generates another latent representation $\mathbf{z}=g(\mathbf{h})$ used for predicting whether two graphs are similar. Given a similar pair $(\mathbf{x}_i,\mathbf{x}_j)$ and a set of graphs $\{\mathbf{x}_k\}$ that are dissimilar from $\mathbf{x}_i$, a contrastive loss~\cite{hadsell2006dimensionality,chen2020simple} can be defined as follows:
\begin{equation}
\label{equ}
    -\log \frac{\textrm{exp}(\textrm{sim}(\mathbf{z}_i, \mathbf{z}_j)/\tau )}{\textrm{exp}(\textrm{sim}(\mathbf{z}_i, \mathbf{z}_j)/\tau )+\sum_{k}\textrm{exp}(\textrm{sim}(\mathbf{z}_i, \mathbf{z}_k)/\tau )}
\end{equation}
where $\textrm{sim}(\cdot,\cdot)$ denotes cosine similarity between two vectors and $\tau$ is a temperature parameter.

Figure~\ref{fig:illustration} presents an illustrative example. Three augmented graphs (AGs) are created from two original graphs (OGs): AG (i) and (ii) are from OG (i); AG (iii) is from OG (ii).  To create AG (i),  three random alteration operations are performed consecutively, including edge deletion, node deletion, and edge insertion. Each operation is applied to the intermediate graph resulting from the last operation. AG (ii) is created by applying node deletion, node insertion, and edge insertion. AG (iii) is created by applying edge deletion, edge insertion, and edge deletion. AG (i) and (ii) are labeled as ``similar" since they originate from the same original graph. AG (ii) and (iii) are labeled as ``dissimilar" since they are created from different original graphs.

\begin{figure}[t]
    \centering
    \includegraphics[width=0.3\textwidth]{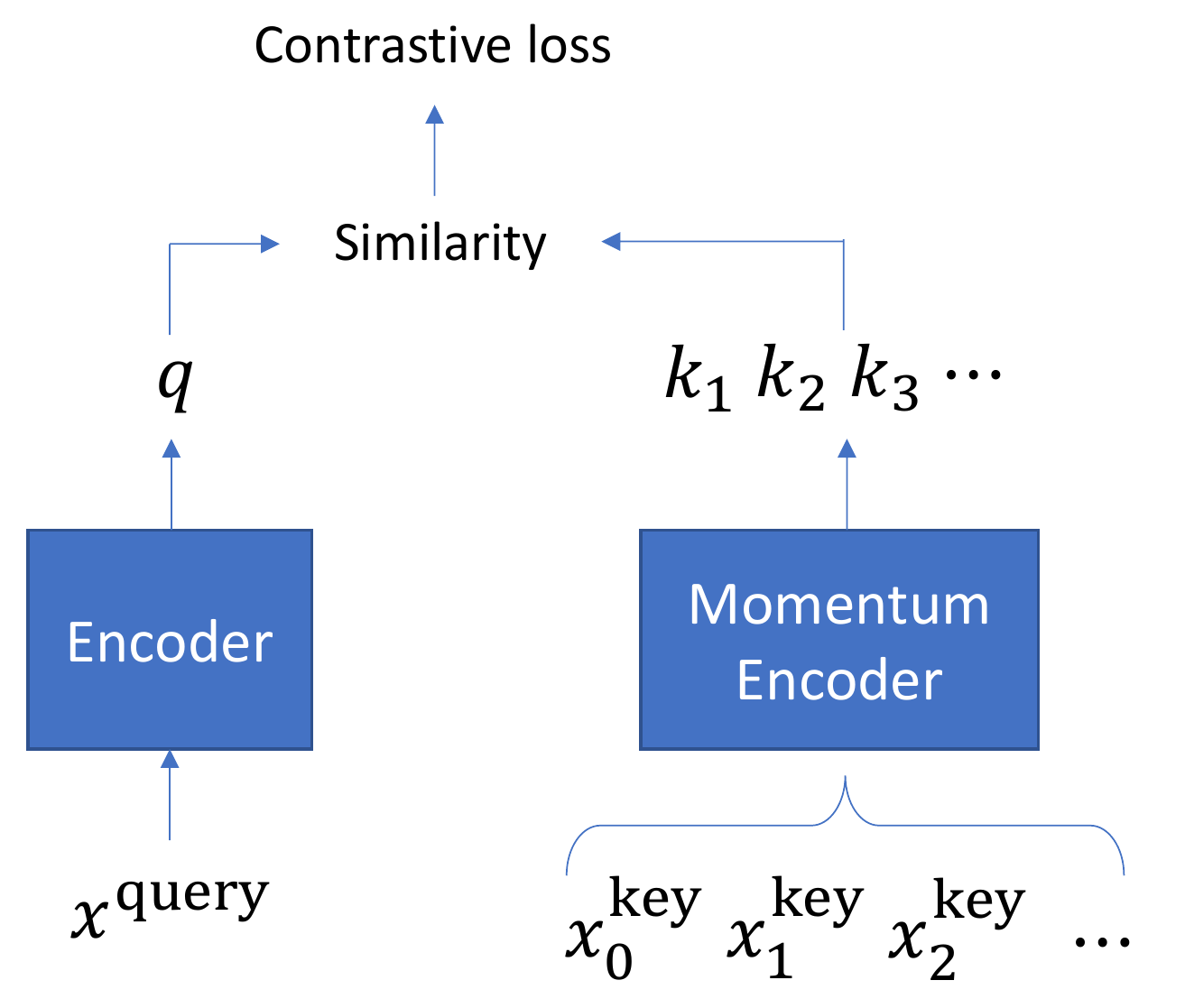}
    \caption{Illustration of MoCo. 
    }
    \label{fig:moco}
\end{figure}

We use MoCo~\cite{he2019moco} to perform efficient optimization of the loss in Eq.(\ref{equ}), based on   a queue that is independent of minibatch size. This queue contains a dynamic set of  augmented graphs (called keys). In each iteration, the latest minibatch of graphs are added into the queue; meanwhile, the oldest minibatch is removed from the queue. In this way, the queue is decoupled from minibatch size. Figure~\ref{fig:moco} shows the architecture of MoCo. The keys are encoded using a momentum encoder. Given an augmented graph (called a query) in the current minibatch and a key in the queue, they are considered as a positive pair if they originate from the same graph, and a negative pair if otherwise.  A similarity score is calculated between the encoding of the query and the encoding of each key. Contrastive losses are defined on the similarity scores and binary labels.

\begin{figure}[t]
    \centering
    \includegraphics[width=0.3\textwidth]{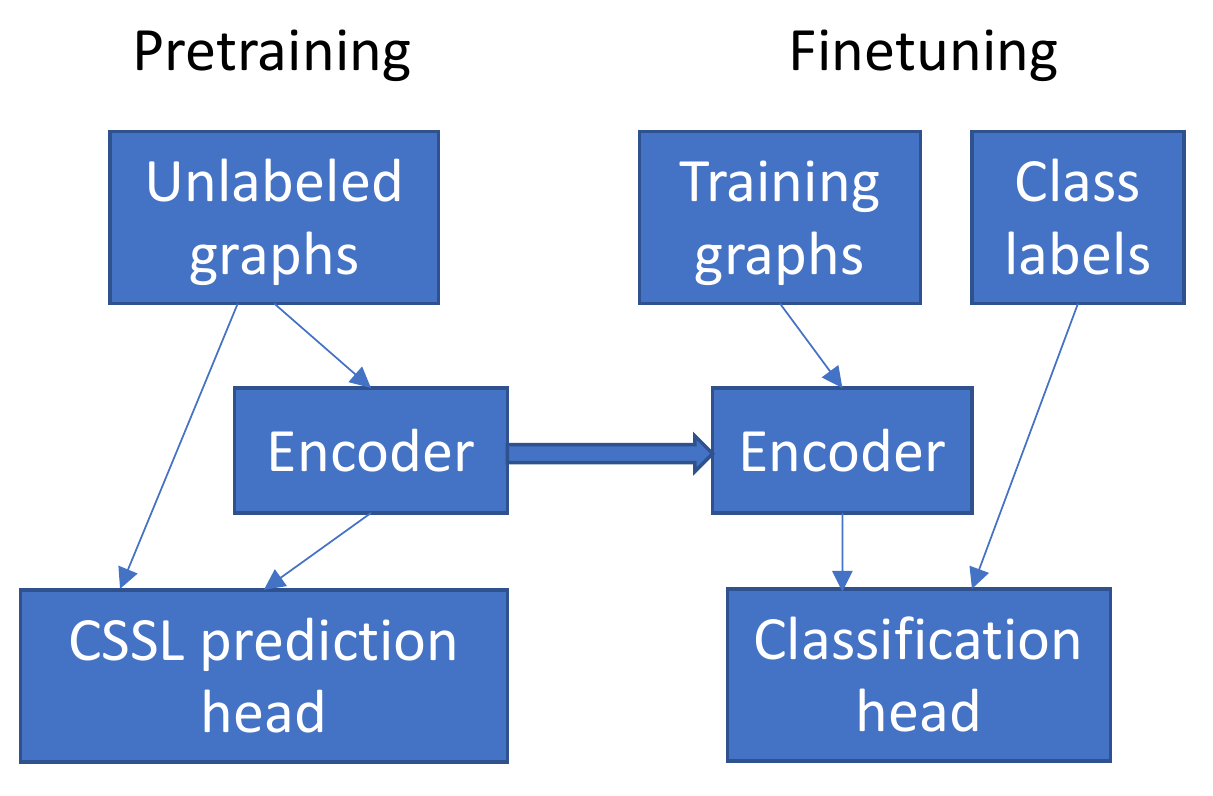}
    \caption{Illustration of CSSL-Pretrain.
    }
    \label{fig:pretrain}
\end{figure}

\begin{figure}[t]
    \centering
    \includegraphics[width=0.3\textwidth]{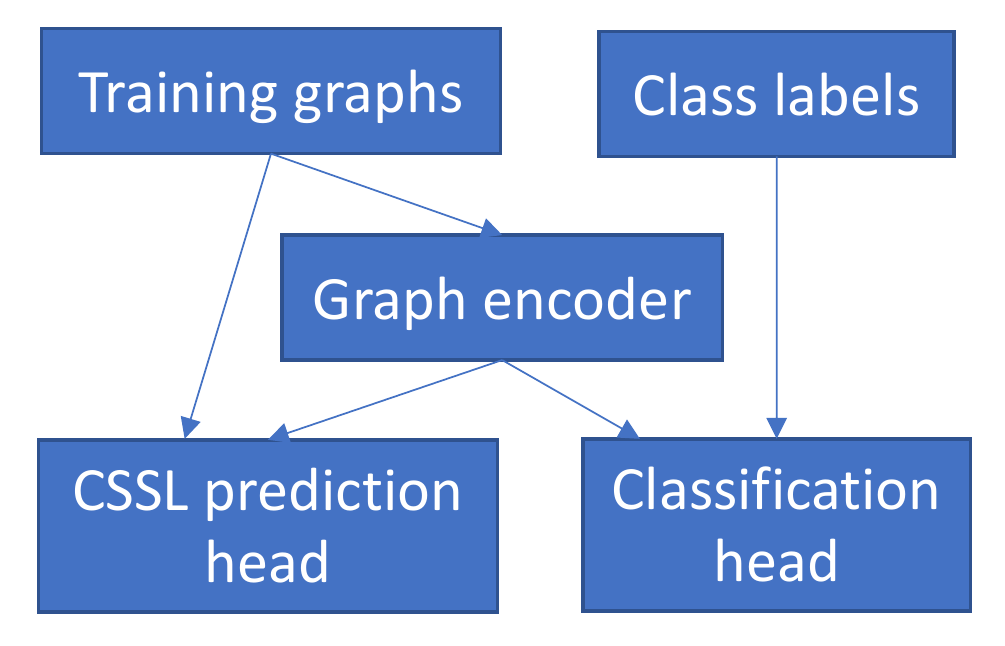}
    \caption{Illustration of CSSL-Reg.
    }
    \label{fig:reg}
\end{figure}

\subsection{CSSL-based Pretraining}
Having presented CSSL on graphs, we study two approaches of using graph CSSL for alleviating overfitting in graph classification. The first approach is to use graph CSSL to pretrain a graph encoder and use this pretrained encoder to initialize the graph classification model. We call this approach CSSL-Pretrain. Given a collection of unlabeled graphs, we define a graph CSSL task on these graphs, then perform this task using a network consisting of a graph encoder and a CSSL-specific prediction head. The head is a multi-layer perceptron which takes graph representations generated by the graph encoder as inputs and predicts whether two augmented graphs are similar. After training, the CSSL-specific prediction head is discarded. Next, we finetune the pretrained graph encoder in the graph classification task. The graph classification network consists of a graph encoder and a classification head. The classification head takes the graph representations generated by the encoder as inputs and predicts the class label. We use the  encoder pretrained by CSSL to initialize the encoder in the classification model and continue to train it on the original graphs and their class labels.


\subsection{CSSL-based Regularization}
The second approach we propose is CSSL-Reg, where we use the graph CSSL task to regularize the graph classification model. Given the training graphs, we encode them using a graph encoder. Then on top of the graph encodings, two tasks are defined. One is the classification task, which takes the encoding of a graph as input and predicts the class label of this graph. The prediction is conducted using a classification head. The other task is graph CSSL. Given the augmented graphs stemming from the  training graphs, CSSL predicts whether two augmented graphs are from the same original graph. The loss of the CSSL task serves as a data-dependent regularizer to alleviate overfitting. The CSSL task has a predictive head. The two tasks share the same graph encoder. Formally, CSSL-Reg solves the following optimization problem:
\begin{equation}
\label{eq:general}
\begin{array}{l}
    \mathcal{L}^{(c)}(D,L;\mb{W}^{(e)}, \mb{W}^{(c)})+ \lambda\mathcal{L}^{(p)}(D,\mb{W}^{(e)},\mb{W}^{(p)} )
    \end{array}
\end{equation}
where $D$ represents the training graphs and $L$ represents their labels. $\mb{W}^{(e)}$, $\mb{W}^{(c)}$, and $\mb{W}^{(p)}$ denote the graph encoder, classification head in the classification task, and prediction head in the CSSL task respectively. $\mathcal{L}^{(c)}$ denotes the classification loss and $\mathcal{L}^{(p)}$ denotes the CSSL loss. $\lambda$ is a tradeoff parameter.



\subsection{Graph Encoder}
At the core of CSSL-Pretrain and CSSL-Reg is to better learn a graph encoder using CSSL. Our methods can be used to learn any graph encoder. 
In this work, we perform the study using the Hierarchical Graph Pooling with Structure Learning (HGP-SL) encoder~\cite{zhang2019hierarchical}, while noting that other graph encoders are also applicable.
HGP-SL is composed of interleaving layers of graph convolution and graph pooling. Graph convolution learns multiple layers of latent embeddings of each node in the graph by leveraging the embeddings of neighboring nodes.
The graph pooling operation selects a subset of informative nodes to form a subgraph.
A node is considered less informative if its representation  can be well reconstructed by those of its neighbors. Given the structure of the pooled subgraph, HGP-SL performs structure learning to refine the structure of the subgraph. HGP-SL calculates the similarity of two nodes in the subgraph and connects them if the similarity score is large enough. Given the refined subgraph, graph convolution and pooling are conducted again. The layers of convolution, pooling, and structure refinement repeat multiple times. A readout function is used to aggregate representations of individual nodes into a single representation of the graph. A multi-layer perceptron serves as the classification head to predict the class label from the graph-level representation.

\begin{table}[t]
      \centering
      \caption{ Statistics of datasets }
      \begin{tabular}{llllll}
        \hline
        Dataset  &  PT$^*$ & D\&D & NCI1 & NCI109 & Mut$^{**}$ \\

        \hline
        \# classes & 2 & 2 & 2 & 2 & 2\\
        \# train& 890 & 942 & 3288 & 3301 & 3469 \\
        \# validation& 111 & 117 & 411 & 412 & 433\\
        \# test & 112 & 119 & 411 & 414 & 435 \\
        Avg. \# nodes & 39.1 & 284.3 & 29.9 & 29.7 & 30.3 \\
        Avg. \# edges & 72.8 & 715.7 & 32.3 & 32.1 & 30.8\\
        \hline
      \end{tabular}
      *PT denotes PROTEINS. **Mut denotes Mutagenicity.
        \label{tab:dataset_statistics}

\end{table}

\begin{table*}[t]
      \centering
      \caption{Graph Classification Accuracy (\%)}
      \resizebox{\textwidth}{!}{
      \begin{tabular}{llcccccc}
        \hline
        Categories & Method  & PROTEINS & D\&D & NCI1 & NCI109 & Mutagenicity \\

        \hline
          Kernels & GRAPHLET  & 72.23$\pm$4.49 & 72.54$\pm$3.83 & 62.48$\pm$2.11 & 60.96$\pm$2.37 & 56.65$\pm$1.74 \\
         & SP  & 75.71$\pm$2.73 & 78.72$\pm$3.89 & 67.44$\pm$2.76 & 67.72$\pm$2.28 & 71.63$\pm$2.19\\
                & WL  & 76.16$\pm$3.99 & 76.44$\pm$2.35 & 76.65$\pm$1.99 & 76.19$\pm$2.45 & 80.32$\pm$1.71\\
        \hline
          GNNs  & GCN  & 75.17$\pm$3.63 & 73.26$\pm$4.46 & 76.29$\pm$1.79 & 75.91$\pm$1.84 & 79.81$\pm$1.58\\
            & GraphSAGE  & 74.01$\pm$4.27 & 75.78$\pm$3.91 & 74.73$\pm$1.34 & 74.17$\pm$2.89 & 78.75$\pm$1.18 \\
                & GAT  & 74.72$\pm$4.01 & 77.30$\pm$3.68 & 74.90$\pm$1.72 & 75.81$\pm$2.68 & 78.89$\pm$2.05\\
        \hline
          Pooling       & Set2Set  & 79.33$\pm$0.84 & 70.83$\pm$0.84 & 69.62$\pm$1.32 & 73.66$\pm$1.69 & 80.84$\pm$0.67 \\
                & DGCNN  & 79.99$\pm$0.44 & 70.06$\pm$1.21 & 74.08$\pm$2.19 & 78.23$\pm$1.31 & 80.41$\pm$1.02 \\
                & DiffPool  & 79.90$\pm$2.95 & 78.61$\pm$1.32 & 77.73$\pm$0.83 & 77.13$\pm$1.49 & 80.78$\pm$1.12\\
         & EigenPool  & 78.84$\pm$1.06 & 78.63$\pm$1.36 & 77.24$\pm$0.96 & 75.99$\pm$1.42 & 80.11$\pm$0.73 \\
                & gPool  & 80.71$\pm$1.75 & 77.02$\pm$1.32 & 76.25$\pm$1.39 & 76.61$\pm$1.39 & 80.30$\pm$1.54\\
                & SAGPool  & 81.72$\pm$2.19 & 78.70$\pm$2.29 & 77.88$\pm$1.59 & 75.74$\pm$1.47 & 79.72$\pm$0.79\\
                & EdgePool  & 82.38$\pm$0.82 & 79.20$\pm$2.61 & 76.56$\pm$1.01 & 79.02$\pm$1.89 & 81.41$\pm$0.88\\
                & HGP-SL  & 84.91$\pm$1.62 & 80.96$\pm$1.26 & 78.45$\pm$0.77 & 80.67$\pm$1.16 & 82.15$\pm$0.58 \\
        \hline
        Self-supervised  & InfoGraph  & 75.18$\pm$0.51 & 74.24$\pm$0.86 & 70.93$\pm$1.78 & 75.70$\pm$1.51 & 72.32$\pm$1.70 \\
                      & GCC-freezing  & 74.48$\pm$3.12 & 75.63$\pm$3.22 &  66.33$\pm$ 2.65 & 66.18$\pm$3.83 & 68.11$\pm$2.78\\
                     & GCC-finetuning  & 69.49$\pm$1.42 & 75.46$\pm$2.44 & 71.00$\pm$1.78 & 69.90$\pm$1.04 & 74.43$\pm$1.35\\

        \hline
        \hline
        CSSL-Freeze & A1-specific  & 84.64$\pm$0.96 & 78.74$\pm$0.92 & 72.60$\pm$1.43 & 76.40$\pm$0.54 & 77.03$\pm$0.66 \\
                  & A1-all & 78.57$\pm$1.64 & 75.96$\pm$1.60 & 72.02$\pm$1.32 & 75.19$\pm$1.00 & 77.08$\pm$0.63 \\
                  & A3-specific  & 80.36$\pm$1.99 & 78.49$\pm$0.94 & 72.70$\pm$1.94 & 76.42$\pm$0.71 & 77.08$\pm$0.48 \\
                  & A3-all & 76.34$\pm$1.92 & 77.73$\pm$1.73 & 71.56$\pm$0.93& 75.70$\pm$1.16& 76.85$\pm$0.84\\
        \hline
         CSSL-Pretrain & A1-specific  & 85.71$\pm$0.69 & 82.02$\pm$1.42 & 78.62$\pm$0.63 & 80.72$\pm$1.06 &  82.00$\pm$0.63\\
         & A1-all & 81.79$\pm$1.50 & 80.84$\pm$1.24 & 78.03$\pm$1.14 & 77.51$\pm$1.37 & 82.23$\pm$0.73\\
        & A3-specific  & 82.77$\pm$1.70 & 80.84$\pm$1.54 & 79.44$\pm$0.67 & 81.01$\pm$1.01 &  82.41$\pm$0.59\\
         & A3-all  & 81.07$\pm$1.63 & 80.25$\pm$1.41 & 78.71$\pm$0.80 & 79.87$\pm$1.06 &  \textbf{82.64$\pm$0.83}\\
        \hline
        CSSL-Reg & A1-specific  & 84.11$\pm$0.87 & \textbf{82.18$\pm$1.34} & 80.07$\pm$0.60 & \textbf{81.16$\pm$1.42} & 82.07$\pm$0.65\\
        & A1-all & 83.57$\pm$1.07 & 80.50$\pm$1.34 & 79.32$\pm$0.75 & 77.80$\pm$1.46 & 80.83$\pm$1.66\\
        & A3-specific & \textbf{85.80$\pm$1.01} & 79.66$\pm$1.71 & \textbf{80.09$\pm$1.07} & 79.69$\pm$1.70 & 81.61$\pm$1.05\\
        & A3-all & 81.61$\pm$1.61 & 79.58$\pm$1.41& 78.64$\pm$0.76 & 79.18$\pm$0.87 & 82.23$\pm$1.04\\
        \hline
      \end{tabular}}
        \label{graph_classification_result}
          ``A1" denotes performing one random graph alteration operation to obtain an augmented graph and ``A3" denotes performing three consecutive random  alteration operations to obtain an augmented graph.  ``specific" denotes using the training graphs in the target dataset to define CSSL losses
          and ``all" denotes using training graphs in all the five datasets to define CSSL losses.
          \\

\end{table*}

\section{Experiments}
\subsection{Dataset}
We used 5 graph classification datasets\footnote{Datasets are publicly available at https://ls11-www.cs.tu-dortmund.de/staff/morris/graphkerneldatasets} in the experiments. Each data example consists of a graph and a class label. In PROTEINS and D\&D, each graph represents a protein. A binary label is associated with each graph, representing whether the protein is a non-enzyme. NCI1 and NCI109 contain graphs representing chemical compounds with labels denoting whether they can inhibit the growth of cancer cells. The graphs in Mutagenicity represent chemical compounds. Each graph is labeled as mutagen or non-mutagen.
We randomly split each dataset into three parts: 80\% for training, 10\% for validation, and 10\% for testing. The random split is repeated for 10 times and the average performance with standard deviation is reported.
The statistics of these datasets are summarized in Table~\ref{tab:dataset_statistics}.

\subsection{Experimental Setup}


\paragraph{CSSL-Pretrain}
For CSSL pretraining,  the queue size in MoCo is set as 1024 for the D\&D and PROTEINS dataset, and 4096 for the NCI1, NCI109, and Mutagenicity dataset. The MoCo momentum is set as 0.999 and the temperature $\tau$ is set as 0.07. The initial learning rate is searched in $\{1e^{-3}, 1e^{-4}, 1e^{-5}\}$ and decayed with the cosine decay schedule \cite{loshchilov2016sgdr}.  We find it beneficial to utilize a small batch size (16 or 32), a small learning rate ($1e^{-5}$), and train for more epochs ($1k \sim 3k$).

For finetuning the classification model, we search the initial learning rate in $\{1e^{-2}, 1e^{-3}, 1e^{-4}\}$ and utilize the Adam optimizer~\cite{kingma2014adam} to optimize the model.  Following \cite{zhang2019hierarchical}, we adopt early stopping based on the validation loss. Specifically, we stop training if the validation loss does not decrease for 100 consecutive epochs. We select the model with the smallest validation loss as the final model.

\subsubsection{CSSL-Reg}
We search the regularization parameter $\lambda$ in $\{1,0.1,0.01,0.001,0.0001\}$. 
The Adam optimizer is used and the initial learning rate is searched in $\{1e^{-2}, 1e^{-3}, 1e^{-4}\}$. We set the queue size in Moco as 512 for the D\&D and PROTEINS dataset, and 2048 for the NCI1, NCI109, and Mutagenicity dataset. The settings of batch size, patience for early stopping, MoCo momentum, and temperature $\tau$ are the same as those in CSSL-Pretrain.

\subsubsection{Graph Encoder} Following~\cite{zhang2019hierarchical}, the dimension of node representation is set to 128. The number of HGP-SL layers is set as 3. The pooling ratio is searched in $[0.1,0.9]$ and the dropout ratio is searched in $[0.0,0.5]$.

\subsection{Baselines} We compare with the following categories of baselines.

\begin{itemize}
    \item

\textbf{Graph Kernel Methods.}  This category of methods compares the similarity of two graphs in a kernel space and performs classification based on the similarity between graphs.
We compare with three  algorithms: GRAPHLET \cite{shervashidze2009efficient}, Shortest-Path (SP) Kernel \cite{borgwardt2005shortest}, and Weisfeiler-Lehman (WL) Kernel \cite{shervashidze2011weisfeiler}.
\item
\textbf{Graph Neural Networks.}
GCN \cite{kipf2016semi}, GraphSAGE \cite{hamilton2017inductive}, and GAT \cite{velivckovic2017graph} are three GNN models  designed for learning node representations in graphs.  Node representations are aggregated into a representation of the entire graph via a readout function and the graph representation is subsequently used for graph classification.
\item
\textbf{Graph Pooling Methods.}
Approaches in this group combine graph neural networks with pooling mechanisms. We compare with eight pooling algorithms, including two global pooling algorithms:  Set2Set \cite{vinyals2015order} and DGCNN \cite{zhang2018end}, and six hierarchical graph pooling methods: DiffPool \cite{ying2018hierarchical}, EigenPool \cite{ma2019graph}, gPool \cite{gao2019graph}, SAGPool \cite{lee2019self}, EdgePool \cite{diehl2019edge}, and HGP-SL \cite{zhang2019hierarchical}.

\item
\textbf{Self-supervised Learning Methods.} We compare with InfoGraph \cite{sun2019infograph}  which maximizes the mutual information between the graph-level representation and the representations of substructures at different scales and GCC \cite{qiu2020gcc} where the SSL task is subgraph instance discrimination.
\item
\textbf{CSSL-Freeze.} We compare with the following setting called CSSL-Freeze. Given a collection of unlabeled graphs, we train the graph encoder using CSSL. Then the graph encoder is directly plugged into the graph classification model without further finetuning. When training the graph classification model, only the classification head is trained and the weights of the graph encoder are frozen.
\end{itemize}

\begin{table}[t]
      \centering
      \caption{L1 difference between training accuracy and testing accuracy. }
      \resizebox{\columnwidth}{!}{
      \begin{tabular}{lccccc}
        \hline
        Datasets  &  PT$^*$ & D\&D & NCI1 & NCI109  & Mut$^{**}$ \\

        \hline
        HGP-SL & 7.4 & 15.6 & 7.8 & 3.6 & 5.2\\
        CSSL-Pretrain & 7.6 & 11.3 & 3.6 & 3.7 & 3.4 \\
        CSSL-Reg & 8.3 & 2.6 & 4.1 & 1.8 &  3.0\\

        \hline
      \end{tabular}
      }
      *PT denotes PROTEINS. **Mut denotes Mutagenicity.
        \label{tab:difference}
\end{table}

\subsection{Results}
The performance on graph classification is reported in Table \ref{graph_classification_result}. From this table, we make the following observations. \textbf{First}, CSSL-Reg and CSSL-Pretrain outperform baseline approaches for graph classification. This demonstrates the effectiveness of our methods in alleviating overfitting. To further confirm this, we measure the difference between accuracy on the training set and test set in Table \ref{tab:difference}. A larger difference implies more overfitting: performing well on the training set and less well on the test set. As can be seen, in most cases, the train-test difference under CSSL-(Pretrain,Reg) is smaller than that under HGP-SL, which demonstrates that our approaches can better alleviate overfitting. CSSL-Pretrain leverages widely-available unlabeled graphs to learn better graph representations that are robust to overfitting. CSSL-Reg encourages the graph encoder to solve an additional task which reduces the risk of overfitting to the data-deficient classification task on the small-sized training data. \textbf{Second}, our methods outperform other self-supervised learning methods in the literature. This is because our methods learn a holistic representation of the entire graph by judging whether two augmented graphs originate from the same graph. To successfully make such a judgment, the encoder needs to capture the global features of the entire graph. However, in baseline SSL methods, self-supervision is performed locally at individual nodes, which loses the global picture on the entire graph. Therefore, the learned representations are not suitable for classifying the entire graph. \textbf{Third}, on 4 out of the 5 datasets, CSSL-Reg performs better than CSSL-Pretrain. In Table \ref{graph_classification_result}, the train-test difference under CSSL-Reg is smaller than that under CSSL-Pretrain, which implies that CSSL-Reg can better prevent overfitting. This is because in CSSL-Reg, the encoder is learned to perform the classification task and CSSL task simultaneously. Thus the encoder is not completely biased to the classification task. In CSSL-Pretrain, the encoder is first learned by performing the CSSL task, then finetuned by performing the classification task. There is a risk that after finetuning, the encoder is largely biased to the classification task on the small-sized training data, which leads to overfitting.  \textbf{Fourth}, performing CSSL on all graphs in the five datasets yields worse accuracy than CSSL on a single target dataset. This is counter-intuitive because it is expected that more data helps to learn better representations in CSSL. One possible reason is that the five datasets have large domain discrepancy. Using graphs from  different domains to pretrain the encoder may render the encoder biased to those domains and eventually generalizes less well on the target domain.  \textbf{Fifth}, CSSL-Pretrain works better than CSSL-Freeze. This is because in CSSL-Pretrain, the encoder is finetuned using the class labels after pretrained using CSSL. The finetuning can make the encoder more discriminative and suitable for solving the classification problem. In CSSL-Freeze, the encoder is not finetuned. As a result, it may not be optimal for the classification task. \textbf{Six}, on 3 out of the 5 datasets, applying three consecutive random operations yields better results than applying one operation only. The reason is that applying three operations makes the augmented graphs more difficult to judge whether they are from the same original graph. Solving a more difficult task makes the learned representations more robust and effective.

\begin{figure}[t]
    \centering
    \includegraphics[width=0.44\textwidth]{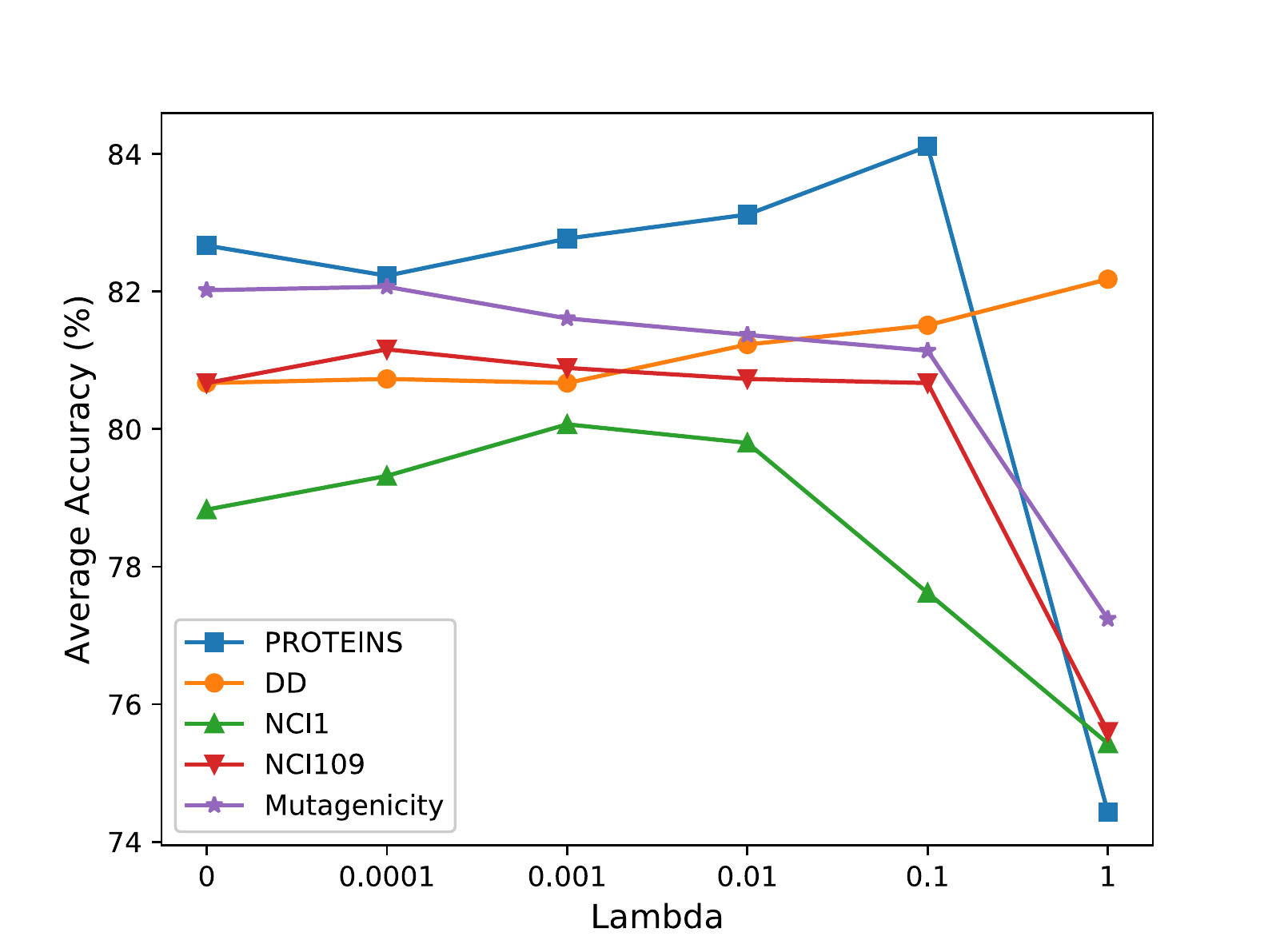}
    \caption{How the regularization parameter in CSSL-Reg affects graph classification accuracy.}
    \label{fig:joint_weight}
\end{figure}

Figure~\ref{fig:joint_weight} shows how the classification accuracy varies as we increase the regularization parameter $\lambda$ in CSSL-Reg. As can be seen, starting from 0, when the regularizer parameter is increasing, the accuracy increases. This is because a larger $\lambda$ imposes a stronger regularization effect, which helps to reduce overfitting. However, if $\lambda$ becomes too large, the accuracy drops. This is because the regularization effect is too strong, which dominates the classification loss.

We also perform a study to verify the importance of randomly selecting graph alteration operations during graph augmentation. We compare with the following deterministic selection setting. For each type of operation including edge insertion, edge deletion, node insertion, and node deletion,  we create augmented graphs by applying this operation once. Table~\ref{tab:operation} shows the average classification accuracy for each operation in CSSL-Reg. As can be seen, the performance of deterministic selection is worse than random selection. The reason is that augmented graphs created by randomly applying alteration operations are more difficult to judge whether they are from the same original graph. Solving a more challenging CSSL task can help to learn representations that are more effective and robust.

\begin{table}[t]
      \centering
      \caption{Performance of CSSL-Reg with deterministic selection of graph alteration operation. }
      \begin{tabular}{lccccc}
        \hline
        Datasets  &  PT$^*$ & D\&D & NCI1 & NCI109  & Mut$^{**}$ \\

        \hline
        Edge Deletion &78.4 & 80.3 & 77.5 & 79.1 & 77.6\\
        Node Deletion & 80.0 & 79.6 & 76.4 & 77.8 &78.4 \\
        Edge Insertion & 78.8 & 80.1 & 76.0 & 75.9 &  81.7\\
        Node Insertion & 77.8 & 79.8 & 78.4 & 76.8 & \textbf{82.1}  \\
      Random & \textbf{84.1}& \textbf{82.2}& \textbf{80.1}& \textbf{81.2}& \textbf{82.1}\\
        \hline
      \end{tabular}
      *PT denotes PROTEINS. **Mut denotes Mutagenicity.
        \label{tab:operation}

\end{table}

\section{Conclusions and Future Works}
In this paper, we propose to use contrastive self-supervised learning to alleviate overfitting in graph classification problems. We propose two approaches based on CSSL. The first approach defines a CSSL task on widely-available  unlabeled graphs and pretrains the graph encoder by solving the CSSL task. The second approach defines a regularizer based on CSSL and the graph encoder is trained to simultaneously minimize the classification loss and the regularizer. We demonstrate the effectiveness of our methods on various graph classification datasets.

For future works, we will extend our approaches to other graph learning problems, such as graph completion, node classification, etc. We will also develop other self-supervised learning methods on graphs, such as predicting which augmented graph is more close to the original graph.





\bibliographystyle{21}
\bibliography{release}

\end{document}